\documentclass[letterpaper,twocolumn,10pt]{article}
\usepackage{usenix2019_v3}

\usepackage{tikz}
\usepackage{amsmath}
\usepackage{biblatex}

\usepackage{multirow}
\usepackage{longtable}
\usepackage{multicol}
\usepackage{caption}
\usepackage{subcaption,graphicx}
\usepackage{dblfloatfix} 
\usepackage{tabularx}
\usepackage{tabulary}
\usepackage{threeparttable}
\usepackage{booktabs}
\usepackage{xcolor}

\newcommand\norm[1]{\left\lVert#1\right\rVert}
\newcommand{\pms}{\sbox0{$1$}\sbox2{$\scriptstyle\pm$}\raise\dimexpr(\ht0-\ht2)/2\relax\box2 }
\newcommand*{\Comb}[2]{{}^{#1}C_{#2}}%
\newcommand{\new}[1]{\textcolor[rgb]{0,0,0}{#1}}

\usepackage{filecontents}

\begin{document}

\date{}

\title{\Large \bf `What's in the box?!': Deflecting Adversarial Attacks by Randomly Deploying Adversarially-Disjoint Models}


\author{
{\rm Sahar Abdelnabi and Mario Fritz}\\
CISPA Helmholtz Center for Information Security
} 

\maketitle

\begin{abstract}
Machine learning models are now widely deployed in real-world applications. However, the existence of adversarial examples has been long considered a real threat to such models. While numerous defenses aiming to improve the robustness have been proposed, many have been shown ineffective. \new{As these vulnerabilities are still nowhere near being eliminated, we propose an alternative deployment-based defense paradigm that goes beyond the traditional white-box and black-box threat models.} Instead of training a single partially-robust model, one could train a set of same-functionality, yet, \textit{adversarially-disjoint} models with minimal in-between attack transferability. These models could then be \textit{randomly and individually} deployed, such that accessing one of them minimally affects the others. Our experiments on CIFAR-10 and a wide range of attacks show that we achieve a significantly lower attack transferability across our disjoint models compared to a baseline of ensemble diversity. In addition, compared to an adversarially trained set, we achieve a higher average robust accuracy while \textit{maintaining} the accuracy of clean examples.
\end{abstract}
\section{Introduction}
Deep neural networks (DNNs) have achieved tremendous success in different tasks (e.g. image recognition~\cite{he2015delving,krizhevsky2012imagenet}, semantic segmentation~\cite{noh2015learning}, and object detection~\cite{redmon2016you}). Besides, they potentially can be used in security-critical applications (e.g. autonomous driving~\cite{tian2018deeptest}, face recognition~\cite{schroff2015facenet}, and phishing detection~\cite{abdelnabi2020visualphishnet}). However, DNNs are vulnerable to \textit{adversarial examples}: inputs with intentionally crafted often-imperceptible noise that can cause misclassification~\cite{goodfellow2014explaining,szegedy2013intriguing,carlini2017towards,kurakin2016adversarial}. These adversarial examples can even happen in the physical world (e.g. photographed images)~\cite{kurakin2016adversarial,kong2020physgan,song2018physical}.

\begin{figure}[!t]
\centering
\includegraphics[width=\linewidth]{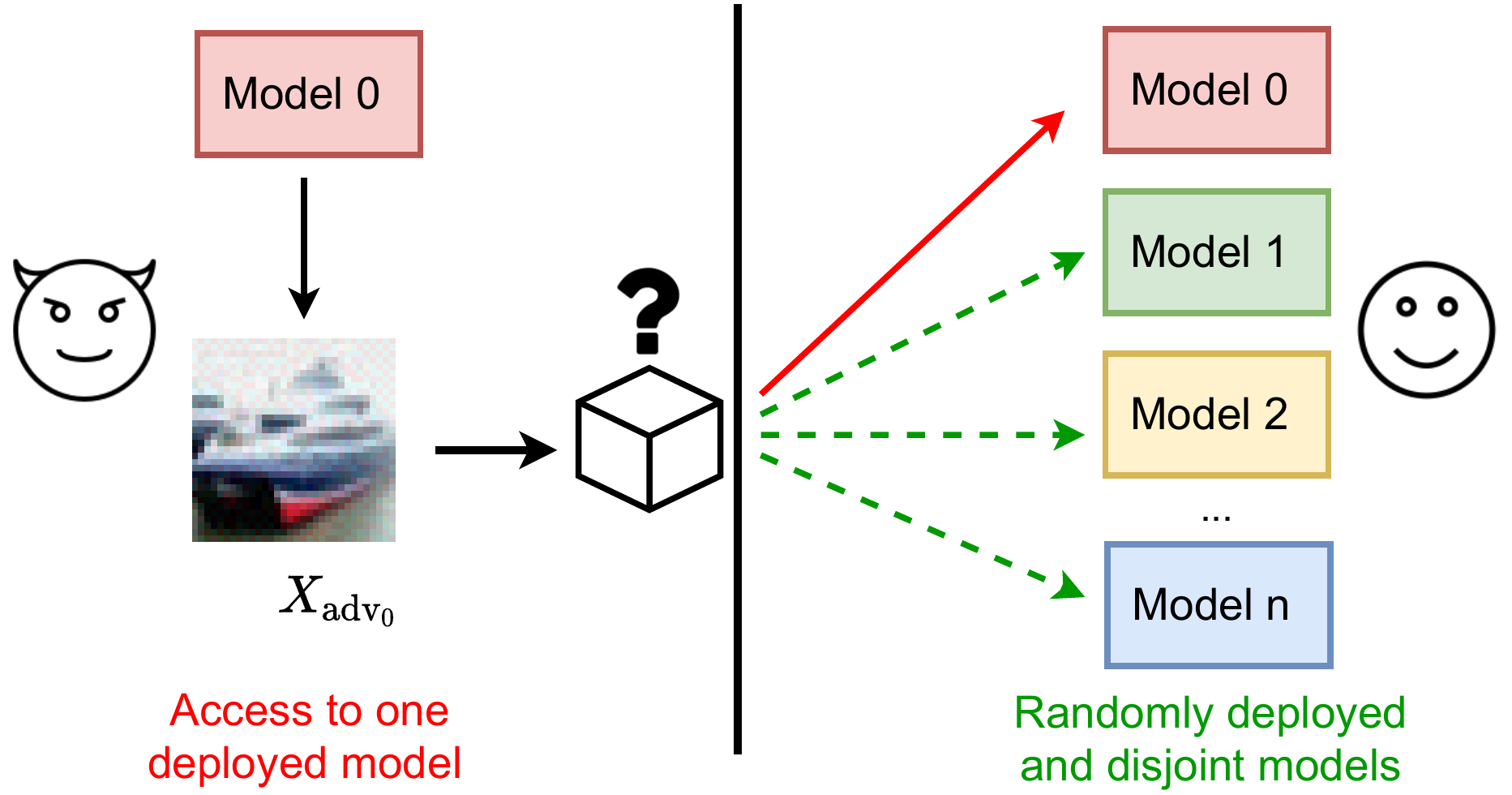}
\caption{We train same-functionality but adversarially-disjoint models with low in-between attack transferability. These models can be deployed individually and randomly as a defense to deflect adversarial attacks.} 
\label{fig:teaser}
\end{figure}

Consequently, adversarial examples raise serious concerns about the security aspects of deploying these models, which fueled an active research line of adversarial attacks and defenses~\cite{akhtar2018threat}. One approach is to make the generation of adversarial examples harder by \textit{gradient obfuscation} (e.g.~\cite{guo2018countering,dhillon2018stochastic,xie2018mitigating}). Another line of defense is to detect adversarial examples at test time~\cite{xu2017feature,meng2017magnet}, however, both were later circumvented by stronger attacks~\cite{athalye2018obfuscated,carlini2017magnet,carlini2017adversarial}. On the other hand, adversarial training~\cite{madry2018towards,goodfellow2014explaining} reduces the sensitivity to adversarial examples by training on them. While being an effective defense against the considered attack, adversarial training is merely a solution as it generalizes poorly to unseen threat models~\cite{stutz2020confidence} while not maintaining the performance on clean examples~\cite{wong2019fast,tsipras2018robustness,madry2018towards,xiao2020one}.  

\paragraph{Disjoint and randomized deployment as a defense.} As it remains unclear whether adversarial examples are escapable and if robustness is within reach~\cite{shafahi2018adversarial,fawzi2018adversarial}, \new{we tackle the problem from a different perspective and propose a deployment-based defense paradigm. We workaround this inherent vulnerability of DNNs: instead of attempting to train and deploy a single model with only so much robustness, we propose to randomly deploy same-functionality, yet, different models. In order for this to succeed, these models should have a minimal in-between transferability of attacks. We call these models \textit{`adversarially-disjoint'}.} \new{By introducing this disjoint randomness in deployment, \textit{we reformulate and go beyond the traditional white-box and black-box threat models.} Even if the adversary can successfully craft adversarial attacks on one model, these attacks are significantly less successful on the other deployed models. Our work is the first to propose a randomized and disjoint deployment strategy for adversarial robustness. We depict an overview of our proposed deployment scenario and disjoint models in~\autoref{fig:teaser}.}

\paragraph{Our approach.} \new{In order to obtain these `adversarially-disjoint' models, we train a set of models jointly for the classification task and minimal transferability. We propose a \textit{novel gradient penalty} that significantly reduces the transferability of attacks across the models. Our approach considerably outperforms a baseline of ensemble diversity and an adversarially trained set. Unlike adversarial training, our approach \textit{does not degrade} the clean examples' accuracy.} 

\paragraph{Comparing with other randomized defenses.} \new{While the concept of introducing randomness as a defense for adversarial attacks was previously introduced~\cite{guo2018countering,dhillon2018stochastic,xie2018mitigating}, it was used as a test time randomness (sometimes non-differentiable) using the same model. This category of defenses is called `gradient masking' or `obfuscated gradients' and were circumvented by stronger attacks (e.g. Expectation over Transformation (EoT) and Backward
Pass Differentiable Approximation (BPDA)~\cite{athalye2018obfuscated}). Unlike these approaches, we do not introduce any randomness at test time and we do not cause any gradient masking; the adversarial attacks can be normally and successfully crafted on one model, but they simply do not transfer well to the other disjoint deployed models. Instead of having adversarial examples that cause the model to fail universally, our approach deflects them such that they mainly cause one random model to fail exclusively. Our approach is similar to gradient masking in making the adversarial examples harder to find, however, we do not obscure the gradient of one model, but we make it less relevant to the defense as a whole.}

\section{Background and Related Work} 
 In this section, we first present a brief overview of adversarial attacks against DNNs, then we summarize existing defenses against these attacks.
 \subsection{Adversarial Examples} \label{ref:adv_related}
 Adversarial examples are malicious input ($x_{\text{adv}}$) that are created by adding a crafted perturbation to a normal image ($x$) which causes a misclassification by a DNN model ($f$) (i.e. $f(x_{\text{adv}})\neq f(x)$)~\cite{dong2018boosting}. There exist many methods for creating adversarial examples by optimizing the added perturbation. We categorize these methods to 1) single-step attacks 2) iterative attacks, and 3) optimization-based attacks.
 
\subsubsection{Single-step Attacks} These methods are optimized for speed as they involve taking only a single step to create the adversarial image. One of the most common single-step methods is the Fast Gradient Sign Method (FGSM)~\cite{goodfellow2014explaining} that finds $x_{\text{adv}}$ by maximizing the training loss function $L(x,y)$, which is usually the cross-entropy. The adversarial examples are created according to:
$$ x_{\text{adv}} = x + \epsilon\cdot\text{sign}(\nabla_x L(x,y)) $$
 
where $\nabla_x L(x,y)$ is the gradient of the loss function w.r.t the input $x$, $y$ is the true label, and $\epsilon$ is the perturbation bound in order to meet the $\ell_\infty$ norm bound (i.e. $\norm{x-x_{\text{adv}}}_\infty \leq \epsilon$).

Similarly, an $\ell_2$ bounded version (i.e. $\norm{x-x_{\text{adv}}}_2 \leq \epsilon$) of FGSM attack is the Fast Gradient Method
(FGM)~\cite{dong2018boosting} which is defined as: $$ x_{\text{adv}} = x + \epsilon\cdot\frac{\nabla_x L(x,y)}{\norm{\nabla_x L(x,y)}_2}$$
 
Another variation of FGSM is random FGSM (R+FGSM)~\cite{tramer2018ensemble} that involves taking a random step ($\alpha$) before adding the gradients, such as: 
$$ x_{\text{adv}} = x^{'} + (\epsilon - \alpha)\cdot\text{sign}(\nabla_{x^{'}} L(x^{'},y))$$ 
where: $x^{'} = x + \alpha\cdot\text{sign}(\mathcal{N}(0^d,1^d)$. R+FGSM was originally used to circumvent gradient masking unintentionally caused by single-step adversarial training.  
 
\subsubsection{Iterative Attacks} Iterative attacks are considered stronger than single-step attacks as they involve taking multiple smaller steps (with a size $\alpha$) by recomputing the gradients at each step. One of the most common iterative attacks is the Projected Gradient Descent (PGD) attack~\cite{madry2018towards} that first takes a random step to form ($x_{\text{adv}_0}$) then iteratively computes: 
$$ x_{\text{adv}_t} = \Pi_{x_\text{adv} \in \Delta_x} [ x_{\text{adv}_{t-1}} + \alpha\cdot\text{sign}(\nabla_x(L(x_{\text{adv}_{t-1}},y) ] $$ where $\Pi_{x_\text{adv} \in \Delta_x}$ is the process of projecting the image back to the allowed perturbation range $\Delta_x$.

To improve the transferability of adversarial examples, Momentum Iterative FGSM (MI-FGSM)~\cite{dong2018boosting} accumulates a velocity vector in
the gradient direction of the loss function across steps. The iterative algorithm is defined as follows:
$$\mathrm{g}_{t+1} = \mu\cdot\mathrm{g}_t + \frac{\nabla_x L(x_{\text{adv}_t},y)}{\norm{L(x_{\text{adv}_t},y)}_1} $$
$$x_{\text{adv}_{t+1}} = x_{\text{adv}_t} + \alpha\cdot\text{sign}(\mathrm{g}_{t+1})$$

where $\mu$ is the decay factor. The intuition of MI-FGSM is that by incorporating the momentum, the optimization is stabilized and can escape local maxima.

\subsubsection{Optimization-based Attacks} Optimization-based attacks directly optimize the distance between the real and adversarial examples in addition to the adversarial loss ($l$). The Carlini and Wagner Attack (CW)~\cite{carlini2017towards} minimizes these two objectives as follows: 
$$ \min \norm{x-x_{\text{adv}}}_2 + c\cdot l(f(x_{\text{adv}}),y_t) $$

where $c$ is a parameter that controls the trade-off between the two objectives, and $y_t$ is the target label. CW is used mostly for $L_2$ norm bounded attacks. 

An extension to CW attack is the Elastic Net attack (EAD)~\cite{chen2018ead}, which uses both $L_1$ and $L_2$ norms in the optimization process: 
$$ \min c\cdot l(f(x_{\text{adv}}),y_t) + \beta\cdot\norm{x-x_{\text{adv}}} + \norm{x-x_{\text{adv}}}_2^2 $$ 
where $\beta$ is also a parameter that controls the trade-off. For both CW and EAD, a parameter ($\kappa$) that controls the misclassification confidence is used in the adversarial loss ($l$) where higher-confidence attacks are more transferable across models. 

\subsection{Defenses against Adversarial Examples}
In this section, we summarize defenses related to our work. 
\subsubsection{Adversarial Training}
Adversarial training is one of the earliest defenses against adversarial attacks which was used in~\cite{goodfellow2014explaining} to train against an FGSM adversary. However, it was later found to be ineffective against a stronger iterative adversary (e.g. PGD)~\cite{madry2018towards}. PGD adversarial training~\cite{madry2018towards} remains empirically robust, unlike several other defenses that were circumvented~\cite{athalye2018obfuscated,wong2019fast}. The robustness in PGD training comes at the cost of more expensive training, e.g. 80 hours are required to train a robust CIFAR-10 model in~\cite{madry2018towards}. This was later improved to 10 hours in~\cite{shafahi2019adversarial} by re-using the gradient in multiple iterations, and to few minutes in~\cite{wong2019fast} by training against an FGSM adversary with a random step and other optimization tricks.

However, it is still an issue that adversarial training causes a drop in the accuracy of clean examples~\cite{shafahi2019adversarial,madry2018towards,wong2019fast,tsipras2018robustness,xiao2020one}. Additionally, it is less effective against unseen threat models~\cite{stutz2020confidence}. In our experiments, we compare the clean examples' accuracy and the average robustness of our set to an adversarially trained set. 

\subsubsection{Ensemble Robustness}
Another defense direction is to use an ensemble of models. The work in~\cite{pang2019improving} aimed to improve the ensemble robustness by promoting diversity in the predictions of each model, such that it would be harder for attacks to succeed in fooling all diverse members in the ensemble. In order to not decrease the accuracy, the maximal prediction of each model should not be affected. Therefore, the diversity regularization works on the non-maximal predictions. 

Our work is closely related to this work as both aim to decrease the attack transferability among the set. However, we directly regularize the gradients of the models w.r.t each other in order to get `adversarially disjoint' models. As we show in the experiments, we achieve a significantly lower transferability across the models compared to this baseline. Additionally, we propose a novel deployment strategy by deploying these disjoint models individually while the work in~\cite{pang2019improving} suggests deploying the whole ensemble.  

\subsubsection{Gradient Masking}
\new{Gradient masking refers to a category of defenses when the gradients are intentionally made harder to be found~\cite{athalye2018obfuscated}. For example, the work in~\cite{guo2018countering} applied random and non-differentiable operations (e.g. cropping, rescaling, and JPEG compression) to the images at test time without changing the model itself. Additionally, the work in~\cite{dhillon2018stochastic} applied a random operation which is Stochastic
Activation Pruning (SAP) that randomly removes a subset of activations. Similarly, the work in~\cite{xie2018mitigating} proposed to apply random operations on the input such as random padding and resizing. These defenses intentionally make the gradients harder to compute and therefore they could cause the optimization of attacks to fail.}

\new{Most of these defenses were circumvented later in~\cite{athalye2018obfuscated} by other adaptive attacks. For example, the non-differentiable operations can be replaced with differentiable approximations in the backward pass, and the randomization can be accounted for using Expectation over Transformation (EoT). It is now a recommended practice to make sure that the defense is not causing gradient masking that can be circumvented~\cite{carlini2019evaluating}.}

\new{Our work is similar to gradient masking or obfuscation in the sense that we intend to make adversarial examples harder to find. However, we do not use any gradient masking for the models. In fact, the white-box attacks for each model individually can be successful. However, they are mostly exclusive to their corresponding model and ideally futile for the others}. 

\subsubsection{Attack Poisoning}
Our work is conceptually similar to previous work that attempts to mitigate attacks by poisoning the attacker's optimization process instead of solving the harder robustness problem. For example, the work in~\cite{shan2020gotta} used `trapdoors' in order to force the attacker's optimization to converge to these trapdoor patterns which could then be detected. However, this could be partially circumvented by stronger adaptive attacks~\cite{carlini2020partial,shan2020gotta} that explicitly avoid the trapdoor signature even if the trapdoors are not known to the adversary. Our work has the advantage that each model in the set is deployed individually and randomly without interacting with the other models and thus, it reveals minimum information about the defense as a whole without needing to obscure the defense mechanism.

Another conceptually similar, yet not related to adversarial defenses, is corrupting the attacker's optimization process in model stealing attacks~\cite{orekondy2019prediction} by poisoning the predictions. The predictions are perturbed such that the adversarial gradient maximally deviates from the original gradient. This is done by maximizing the angular deviation between the two. In our work, we also utilize angular deviation regularization, but we found it inadequate to achieve low transferability between the models, and therefore we propose novel and explicit transferability regularization losses.

\section{Threat Model} \label{ref:threat_model}
\new{Our proposed deployment-based defense and `adversarially-disjoint' models reframe the traditional white-box and black-box threat models. Traditionally, breaking a deployed model (using black-box or white-box attacks) translates to effectively breaking all models. In our work, we workaround this and ask whether there is there a way to alleviate the adversarial attacks through a smarter deployment strategy. Our `adversarially-disjoint' approach is the first step for deployment-based defenses that exploit possible randomization opportunities.} 

\paragraph{Assumptions about the adversary.} We assume that the `disjoint models' would be deployed individually and randomly. We assume that the models are either released randomly from the beginning or released gradually and adaptively in a staged-release as a `zero-day' defense if a model becomes frequently attackable. Therefore, in the normal setup, we assume that the adversary has white-box access to one of the models in the set. We assume that they do not have access to the training data (as commonly adopted in previous work~\cite{shan2020gotta,papernot2016distillation}). We define three types of adversaries according to their knowledge about the models:
\begin{enumerate}
    \item Static adversary: they have no knowledge that different models are deployed, and therefore perform normal attacks without adaptation. 
    \item Skilled adversary: they know that different models are deployed, but have access to one model only, and therefore they attempt to craft highly transferable adversarial examples such that they would transfer to the other deployed models.  
    \item Oracle adversary: they have access to one or more models and they use them to craft ensemble white-box attacks. 
\end{enumerate}

\paragraph{Requirements.} We set two requirements for our models that we measure in our experiments: 1) the classification accuracy of clean examples should be minimally affected by training the disjoint models. 2) Attacks should be minimally transferable from one model to another. 

\section{Adversarially-Disjoint Models}
\begin{figure*}[!t]
\centering
\includegraphics[width=\linewidth]{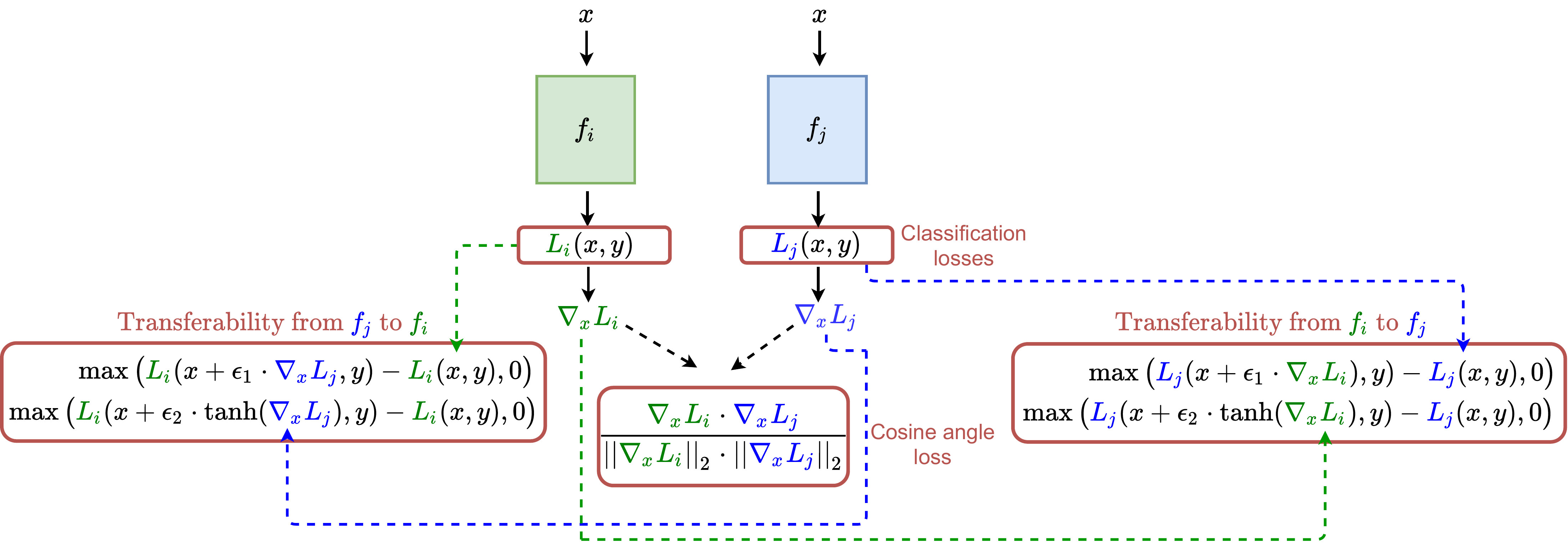}
\caption{Overview of training the adversarially-disjoints sets. Boxes in red indicate training losses.}
\label{fig:method}
\end{figure*}

In this section, we describe our models' set optimization process which is depicted in~\autoref{fig:method} for a pair of models in the set. We train the models jointly to optimize the classification loss and minimize the transferability of attacks.

\paragraph{Classification.} Each model in the set is trained individually for the classification task. The total classification loss is the sum of the classification losses of all models: 
\begin{equation}
\text{L}_{\text{class}} = \sum_{i}^{n}L_i(x,y) 
\end{equation}
where $L_i$ is the categorical cross-entropy loss of model $f_i$ and $n$ is the total number of models.

\paragraph{Gradient angular deviation.} As in previous work~\cite{orekondy2019prediction,kariyappa2019improving,jalwana2020orthogonal}, increasing the angular deviation or misalignment between gradients helps to make the losses uncorrelated which contributes to less transferability of attacks~\cite{kariyappa2019improving,jalwana2020orthogonal}. To increase the angular deviation, we minimize the cosine of the angles between the pairwise gradients in the set:
\begin{equation}
\text{L}_{\text{angle}} = \sum_{i=0}^{n} \sum_{j=i+1}^{n} \frac{\nabla_x L_i\cdot\nabla_x L_j} { \norm{\nabla_x L_i}_2 \cdot \norm{\nabla_x L_j}_2}
\end{equation}

where $\nabla_x L_i$ is the adversarial gradient of model $f_i$. 

\paragraph{Transferability losses.} Increasing the angular deviation helps to decrease the transferability across models. However, in our experiments, we found it less helpful when training more than two models. The core idea of our approach is to explicitly penalize the transferability of attacks between models, and thus, we propose these novel transferability losses. The intuition is: the white-box examples of model $f_j$ (i.e. computed using the adversarial gradients $\nabla_x L_j$) should not be transferable to model $f_i$ (i.e. causing misclassification by increasing the loss of $f_i$). Therefore we minimize: 
\begin{equation} \label{eqn:transfer1}
\text{L}_\text{transfer1} = \sum_{i}^{n} \sum_{j\neq i}^{n} \max \big(L_i(x+\epsilon_1\cdot\nabla_x L_j,y)-L_i(x,y),0\big)
\end{equation}

where $\epsilon_1$ controls the perturbations range and the loss is computed for all pairwise combinations (excluding $i=j$). This means that we penalize the increase of the loss of model $f_i$ incurred by adding the gradients of model $f_j$.

To extend the transferability minimization to $\ell_\infty$ attacks (i.e. sign of the gradient), we approximate the non-differentiable sign operation with the differentiable `$\tanh$' function and adapt the previous loss as follows: 
\begin{equation} \label{eqn:transfer2}
\text{L}_\text{transfer2} = \sum_{i}^{n} \sum_{j\neq i}^{n} \max(L_i(x+\epsilon_2\cdot\tanh(\nabla_x L_j),y)-L_i(x,y),0)
\end{equation}

Similarly, $\epsilon_2$ controls the perturbations range. The models are then trained end-to-end with weighted averaging of the previous losses. 
\begin{equation} \label{eqn:all}
\text{L}_\text{total} = w_1\cdot\text{L}_{\text{class}} + w_2\cdot\text{L}_{\text{angle}} + w_3\cdot\text{L}_\text{transfer1} + w_4\cdot\text{L}_\text{transfer2}
\end{equation}

As a general intuition, the objective of the transferability losses is to search for an adversarial gradient for each model that minimally affects the other models and to make each model ideally exclusively sensitive to its own gradients. 

\section{Experimental Results}
\begin{table*}[!t]
    \centering
    \resizebox{0.85\linewidth}{!}{
    \begin{tabular}{r|l} \toprule
       \textbf{Attack}  &  \textbf{Attack's parameters}\\ \midrule
        FGSM   & $\epsilon=0.031$ \\
        FGM & $\epsilon=1$ \\
        R+FGSM &  $\epsilon=0.031$, $\alpha=\frac{\epsilon}{2}$ \\ \midrule
        $\text{PGD}_1$ & $\epsilon=0.031$, $\alpha=0.0078$, steps=7 \\
        $\text{PGD}_2$ & $\epsilon=0.031$ $\alpha=0.0078$, steps=20 \\
        $\text{MI-FGSM}_1$ & $\epsilon=0.031$, $\alpha=0.0031$, $\mu=1$, steps=10 \\ 
        $\text{MI-FGSM}_2$ & $\epsilon=0.031$, $\alpha=0.0031$, $\mu=1$, steps=20 \\ \midrule
        $\text{CW}_1$ & $c=1.0$, $\kappa=0$, max iterations = 1000, learning rate=0.01, optimizer=Adam \\
        $\text{CW}_2$  & $c=1.0$, $\kappa=40$, max iterations = 1000, learning rate=0.01, optimizer=Adam  \\
        $\text{EAD}_1$ & $c=20.0$, $\kappa=0$, $\beta=0.01$, decision rule=`EN', max iterations = 1000, learning rate=0.01, optimizer=SGD \\
        $\text{EAD}_2$ & $c=10.0$, $\kappa=55$, $\beta=0.01$, decision rule=`EN', max iterations = 1000, learning rate=0.01, optimizer=SGD  \\ \bottomrule
    \end{tabular}}
    \caption{The attacks we use in our experiments and their parameters. The subscripts used here are used to differentiate between different parameters' settings of the same attack and used to refer to these settings in the rest of the paper.}
    \label{tab:attacks_params}
\end{table*}

In this section, we first present the implementation details and our experimental setup. We then evaluate the robustness of our models; we demonstrate the transferability across the adversarially-disjoint models in comparison with baselines. Second, we demonstrate the effective deployed accuracy by considering the accuracy of the whole set. We then evaluate an advanced attack where the adversary can have access to more than one model in the set. Finally, we present an ablation study that shows the effect of different losses and design choices.   

\subsection{Implementation Details}
We used the CIFAR-10 dataset~\cite{krizhevsky2009learning} as one of the most commonly used datasets for studying adversarial robustness (e.g.~\cite{shan2020gotta,pang2019improving,shafahi2019adversarial,xiao2020one,wong2019fast,madry2018towards}). It consists of 50k color training images and 10k colored testing images spanning 10 classes. All experiments are done using the PreAct ResNet18 architecture~\cite{he2016identity,wong2019fast}. We used the PyTorch framework\footnote{https://pytorch.org/} for all our experiments. We varied the number of models in the set from 3 to 8. For training 3 and 4 models, we use all combinations of pairwise transferability losses in each iteration. However, starting from 5 models, we found it more helpful (for convergence, faster training, and memory) to stochastically sample 3 different models at each iteration. 
We used the SGD optimizer with a momentum of 0.9 and weight decay of $1*10^{-4}$. We used the cyclic learning rate~\cite{smith2017cyclical,wong2019fast} where the learning rate increases linearly from 0 to 0.2 during the first half of iterations and then decreases linearly. We set the values of $\epsilon_1$ and $\epsilon_2$ in~\autoref{eqn:transfer1} and~\autoref{eqn:transfer2} to 6 and 0.031 respectively. We trained the models for 75 epochs in the case of 3 and 4 models, 100 epochs in the case of 5 models, 120 epochs in the case of 6 models, and 130 epochs for 7 and 8 models. The batch size ranges from 128 (for 3 models) to 75 (for 8 models). For each loss type (classification, angular deviation, and transferability), we average the losses across all combinations. For the weights in~\autoref{eqn:all}, we set the classification weight to 1 and the other losses' weights to 0.5, training was not sensitive to the exact values of these weights. We will make our code, setup, and models' checkpoints available at the time of publication. 

\subsection{Experimental Setup}
\paragraph{Metrics.} We evaluate the models by the clean test accuracy. We form a transferability matrix of attacks across the models (i.e. an $n\times n$ matrix where the source models are the rows and the target models are the columns). We evaluate the accuracy of black-box attacks between the models in the set (i.e. the non-diagonal elements) and the accuracy of the whole set across all combinations (i.e. including the diagonal white-box elements when the source and target models are the same).

\paragraph{Baselines.} We evaluate our approach against two baselines: 1) the ensemble diversity in~\cite{pang2019improving} where we compare the transferability across our adversarially-disjoint models to their diverse ensemble. 2) An adversarially trained set, as an even more challenging setup than a single adversarially trained model. We compare our approach with adversarial training in terms of the transferability and the average accuracy across the whole set (for an objective and fair comparison, since our models have low transferability but on the other hand no white-box robustness).

We followed~\cite{wong2019fast}'s approach in fast adversarial training. We trained models separately with different random initialization. As in~\cite{wong2019fast}, the models are trained using random step+FGSM and cyclic learning rate. Each model was trained for 50 epochs. We reached a comparable clean and robust accuracy to the PGD training in~\cite{madry2018towards} (e.g. we reached a clean accuracy of 85.2\% versus 87.3\%, and a robust accuracy of 49.53\% versus 50\% for PGD attack with $\epsilon=0.031$, $\alpha=0.0078$, and 7 steps). To train the diverse ensemble~\cite{pang2019improving}, we used the authors' implementation\footnote{https://github.com/P2333/Adaptive-Diversity-Promoting}.
\begin{table*}[!t]
    \centering
    \setlength\tabcolsep{14pt}
    \resizebox{0.9\linewidth}{!}{
    \begin{tabular}{c |c |c c c c c c} \toprule 
         \textbf{Attack} & \textbf{AT} & \multicolumn{6}{c}{\textbf{Adversarially-disjoint models (ours)}} \\
         &     & \textbf{3 models} & \textbf{4 models} & \textbf{5 models} & \textbf{6 models} & \textbf{7 models} & \textbf{8 models} \\ \midrule
         FGSM & 63.1\pms0.21 & 98.6\pms0.55 & 97.0\pms1.85 & 98.3\pms0.85 & 98.0\pms0.85 & 97.9\pms0.92 & 97.5\pms0.97 \\ 
         FGM & 48.3\pms0.25 & 97.7\pms0.89& 95.0\pms3.42 & 97.5\pms1.37 & 97.5\pms1.43 & 97.4\pms1.47 & 97.1\pms1.92 \\
         R+FGSM & 74.9\pms0.24& 95.8\pms0.90& 94.2\pms1.84 & 95.4\pms1.33 & 94.9\pms1.33 & 94.9\pms1.25 & 94.2\pms1.18 \\ \midrule
         $\text{PGD}_1$ & 60.8\pms0.11 & 96.2\pms0.99 & 93.8\pms2.07& 95.1\pms1.95 & 93.9\pms2.92 & 92.8\pms3.54 & 92.1\pms4.95 \\
         $\text{PGD}_2$ & 59.0\pms0.14 & 95.8\pms0.94 & 92.7\pms2.50 & 93.8\pms2.59 & 92.3\pms4.35 & 91.1\pms5.63 & 89.9\pms7.91 \\
         $\text{MI-FGSM}_1$ & 61.4\pms0.10& 98.2\pms0.64 & 95.8\pms2.33 & 96.5\pms1.77 & 95.8\pms1.86 & 95.0\pms2.28 & 94.5\pms2.38 \\ 
         $\text{MI-FGSM}_2$ & 34.4\pms0.50 & 96.2\pms1.33 & 91.1\pms4.95 & 91.8\pms3.91 & 88.5\pms5.87 & 85.4\pms10.18 & 83.7\pms11.6 \\ \midrule
         $\text{CW}_1$  & 74.1\pms0.77 & 93.6\pms0.33& 93.1\pms0.45& 93.3\pms0.53 & 93.3\pms0.36 & 93.5\pms0.58 & 93.2\pms0.69  \\
         $\text{CW}_2$  & 8.3\pms0.83 & 89.5\pms2.14 & 84.3\pms4.34 & 79.7\pms10.12 & 78.3\pms11.11 & 82.9\pms12.92 & 79.1\pms18.61 \\
         $\text{EAD}_1$ & 58.8\pms0.87& 88.8\pms0.48 & 88.2\pms1.55 & 88.7\pms1.91 & 90.1\pms1.65 & 90.6\pms1.93 & 90.7\pms1.54\\
         $\text{EAD}_2$ & 13.2\pms1.34 & 80.8\pms6.64 & 78.4\pms6.43 & 83.2\pms5.17 & 81.5\pms7.16 & 90.3\pms4.91 & 82.8\pms11.93 \\ \bottomrule
    \end{tabular}}
    \caption{Test accuracy (\%) of black-box adversarial examples in the case of adversarially trained (AT) models (3 models trained individually with random seeds), and our approach of adversarially-disjoint models with a varying number of models in the set. We show the average and standard deviation of all black-box combinations. The attacks' parameters are in~\autoref{tab:attacks_params}.}
    \label{tab:transerability}
\end{table*}

\paragraph{Attacks.} We evaluate the models using the range of attacks discussed in~\autoref{ref:adv_related}. Since optimization-based attacks do not explicitly use the gradients of the training loss, we use them to test the generalization of our approach to unseen attacks. In order to evaluate against a skilled adversary (discussed in~\autoref{ref:threat_model}), we evaluate against highly transferable attacks such as high confidence CW and EAD, and MI-FGSM attacks. We also opt to evaluate against attacks that introduce randomness before gradient steps (such as R+FGSM and PGD) since the models were trained without adding a random step first and therefore these attacks are more challenging to our models than their counterparts (FGSM and I-FGSM). We show the attacks' settings and parameters which we use in our experiments in~\autoref{tab:attacks_params}.

\subsection{Transferability among Models}
In this section, we show our evaluations of attack transferability between the adversarially-disjoint models. 

\paragraph{Comparing to adversarial training.}
In~\autoref{tab:transerability}, we show the black-box accuracy of attacks across the set in the case of our approach in comparison with an adversarially trained set. We compute the average accuracy over all black-box combinations of having different source and target models (i.e. the average of the non-diagonal elements in the $n\times n$ transferability matrix). To study the effect of increasing the number of models, we show the average accuracy in the case of training 3 to 8 models. We evaluate the models against single step attacks, iterative attacks, and optimization-based attacks. We include strong and high-confidence optimization-based attacks and highly transferable MI-FGSM with a larger number of steps and a larger step size than what was originally used in~\cite{dong2018boosting}. 

From the table, we highlight the following conclusions: 
\begin{enumerate}
    \item Our approach is significantly more resilient to transferability across all attacks and the number of models in the set, even with highly transferable attacks that transfer well in the adversarially trained set.
    \item As discussed earlier, the transferability of R+FGSM is slightly higher than FGSM, although it is generally a weaker attack as indicated by the adversarially trained set. The random step done in R+FGSM may account for this result. 
    \item The black-box accuracy of single-step attacks is consistently very high and it almost does not deteriorate with increasing the number of models.
    \item When increasing the number of models, the black-box accuracy of iterative attacks ($\text{MI-FGSM}_2$ and PGD) and highly transferable optimization-based attacks gradually declines. 
\end{enumerate}
 
Therefore, in order to evaluate whether it is beneficial to increase the number of models beyond a certain limit (e.g. from 7 to 8 models), we evaluate in the next sections the effective average accuracy of the whole set (i.e. the average of the $n\times n$ matrix including white-box and black-box pairs) and advanced attacks crafted using multiple models. 

\begin{table}[!b]
    \centering
    \resizebox{\linewidth}{!}{
    \begin{tabular}{c|c|c} \toprule 
     \textbf{Attacks} & \textbf{Diverse ensemble} & \textbf{Adversarially-disjoint models (ours)} \\ 
      \midrule
     FGSM & 60.0\pms1.08 & 98.5\pms0.55 \\ 
     $\text{PGD}_1$ & 58.8\pms2.01 & 96.2\pms0.99\\ 
     $\text{PGD}_2$ & 51.1\pms2.69 & 95.7\pms0.94\\
     $\text{MI-FGSM}_1$ & 54.8\pms2.23 & 98.2\pms0.64 \\ 
     $\text{MI-FGSM}_2$ & 47.4\pms2.30 & 96.2\pms1.33 \\
     \bottomrule
    \end{tabular}}
    \caption{Test accuracy (\%) of black-box adversarial examples in the case of diverse ensemble baseline~\cite{pang2019improving} and our approach of adversarially-disjoint models, both are sets of 3 models. We show the average and standard deviation of all black-box combinations. The attacks' parameters are in~\autoref{tab:attacks_params}.}
    \label{tab:diverse_ens}
\end{table}

\paragraph{Comparing to ensemble diversity.}
We next compare our approach to the diverse ensemble in terms of transferability among the set, which we show in~\autoref{tab:diverse_ens} using a representative set of attacks. The experiments are done using 3 models in the set for both approaches. Similar to~\autoref{tab:transerability}, we report the average black-box accuracy across all black-box combinations in the set. As can be observed, the black-box accuracy in our approach significantly outperforms the ensemble diversity. In our work, we propose deploying the models individually in order to introduce uncertainty for the adversary, and therefore, we focus on minimizing the transferability. On the other hand, the work in~\cite{pang2019improving} proposed to deploy the ensemble as a whole and, therefore, the gained robustness was mainly in the white-box case on the ensemble instead of the black-box case across the models. However, the white-box robustness gained by the diverse ensemble was significantly reduced by stronger attacks in a subsequent work~\cite{tramer2020adaptive}. 
\begin{figure} [!b]
    \centering
    \includegraphics[width=0.9\linewidth]{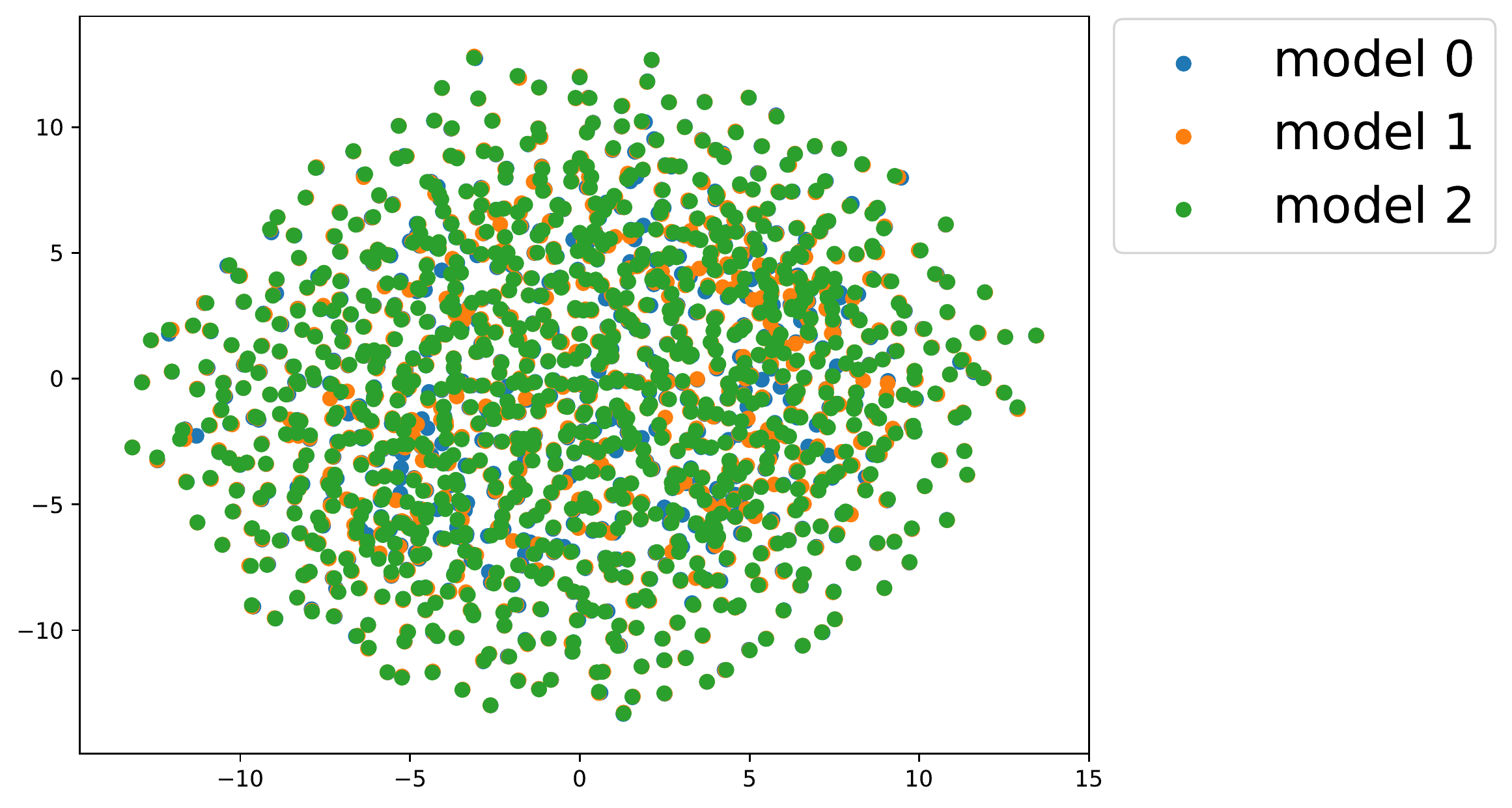}
    \caption{Visualizing the adversarial gradients' signs of the adversarially trained models using t-SNE embeddings for 1000 images.}
    \label{fig:tsne_at}
\end{figure}
\begin{figure*}[!t]
\centering
\begin{subfigure}{0.45\linewidth}
  \centering
  \includegraphics[width=\linewidth]{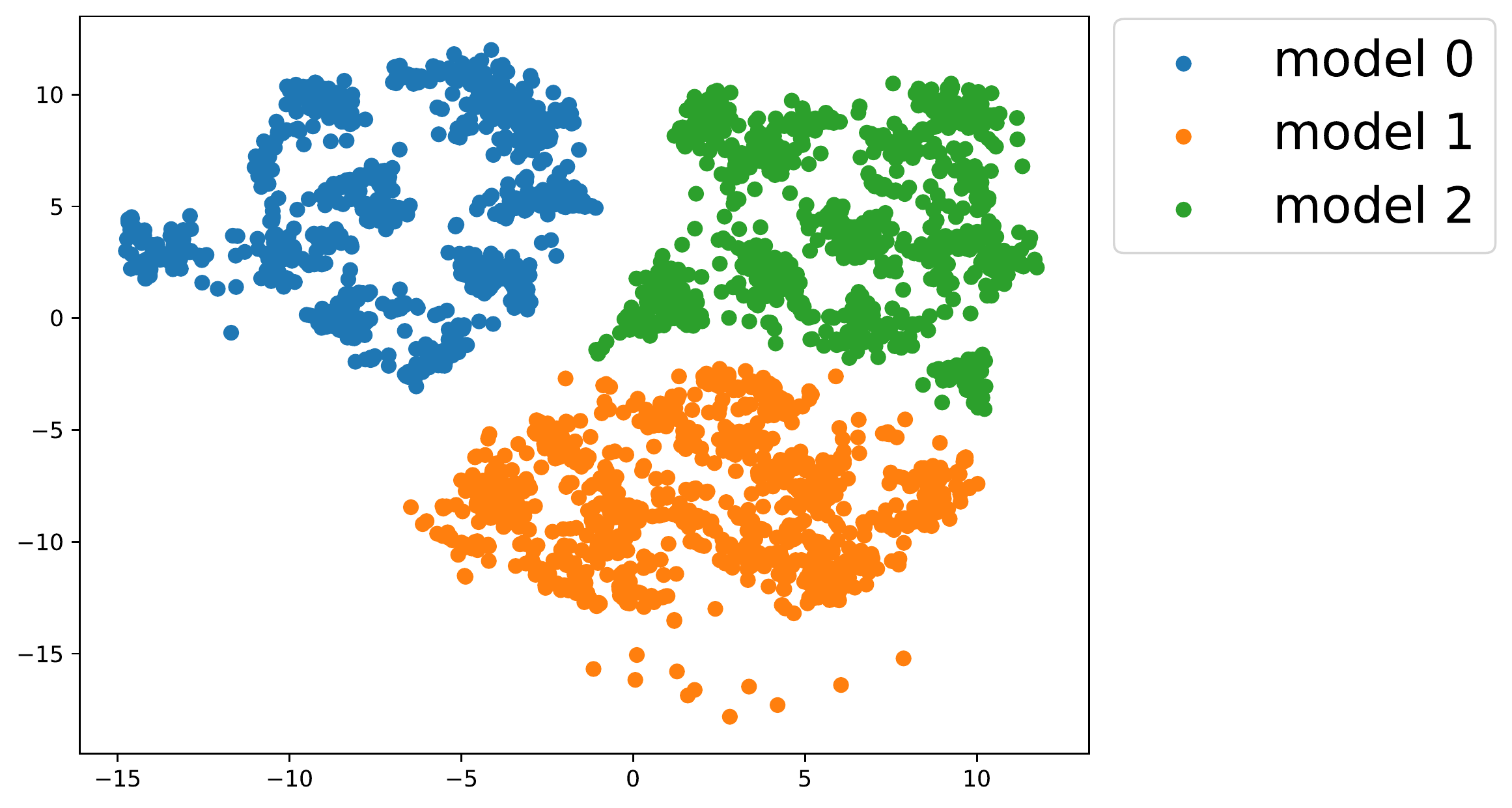}  
  \caption{Adversarially-disjoint (3 models).}
  \label{fig:ours_tsne}
 \end{subfigure}
\begin{subfigure}{0.45\linewidth}
  \centering
  \includegraphics[width=\linewidth]{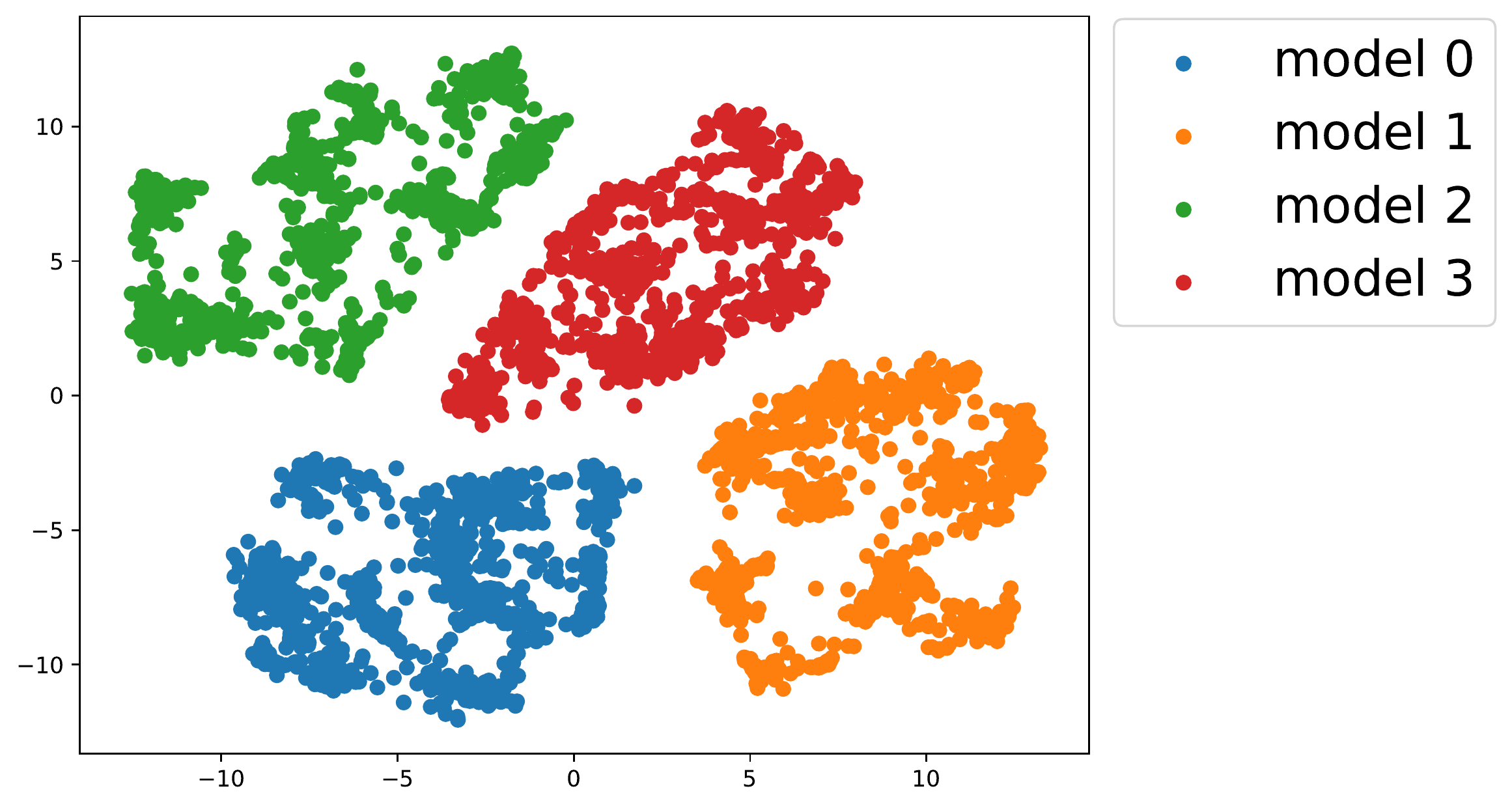}  
  \caption{Adversarially-disjoint (4 models).}
  \label{fig:ours_tsne2}
 \end{subfigure}

 \begin{subfigure}{0.45\linewidth}
  \centering
  \includegraphics[width=\linewidth]{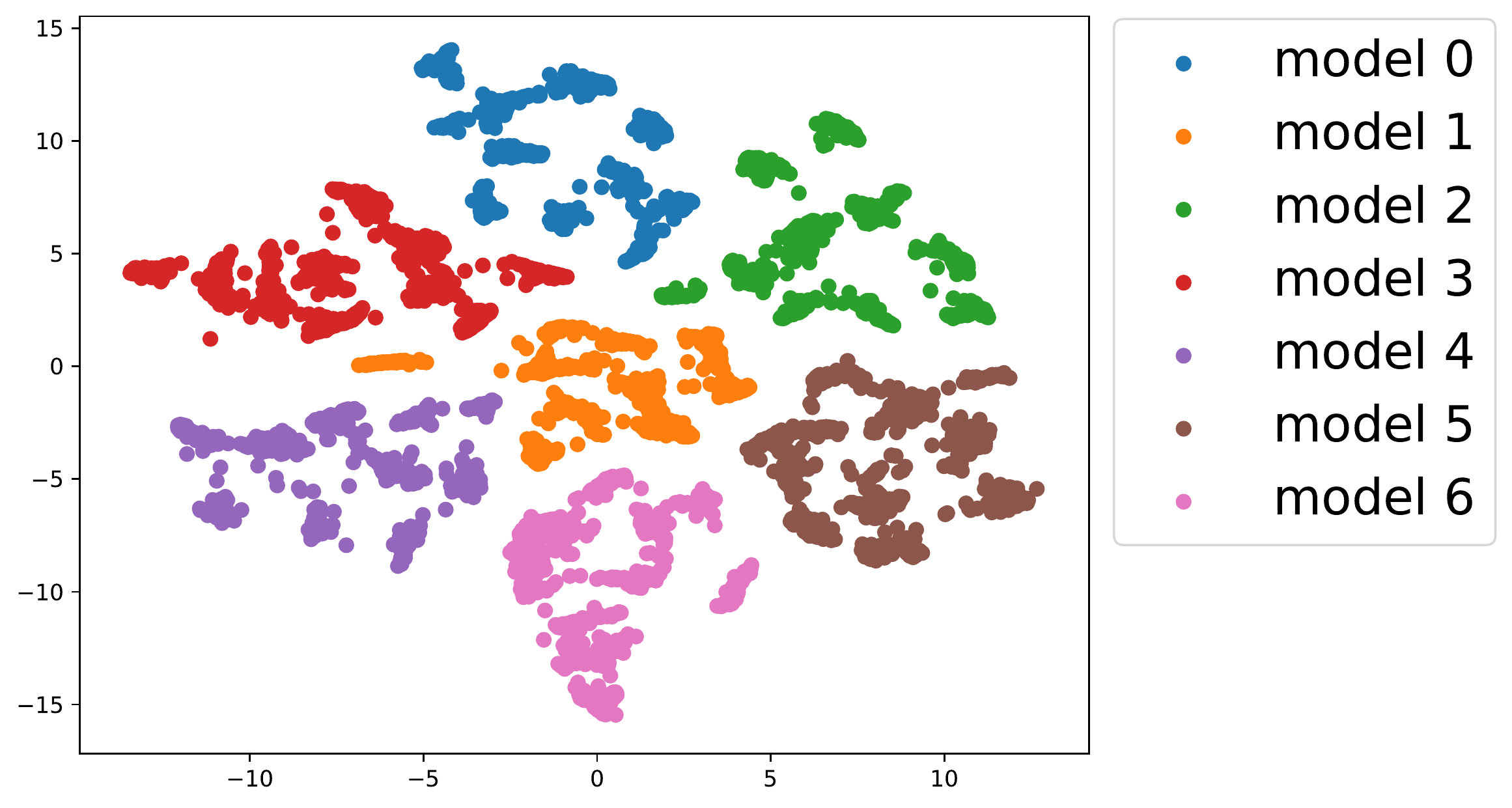}  
  \caption{Adversarially-disjoint (7 models).}
  \label{fig:ours_tsne3}
\end{subfigure}
 \begin{subfigure}{0.45\linewidth}
  \centering
  \includegraphics[width=\linewidth]{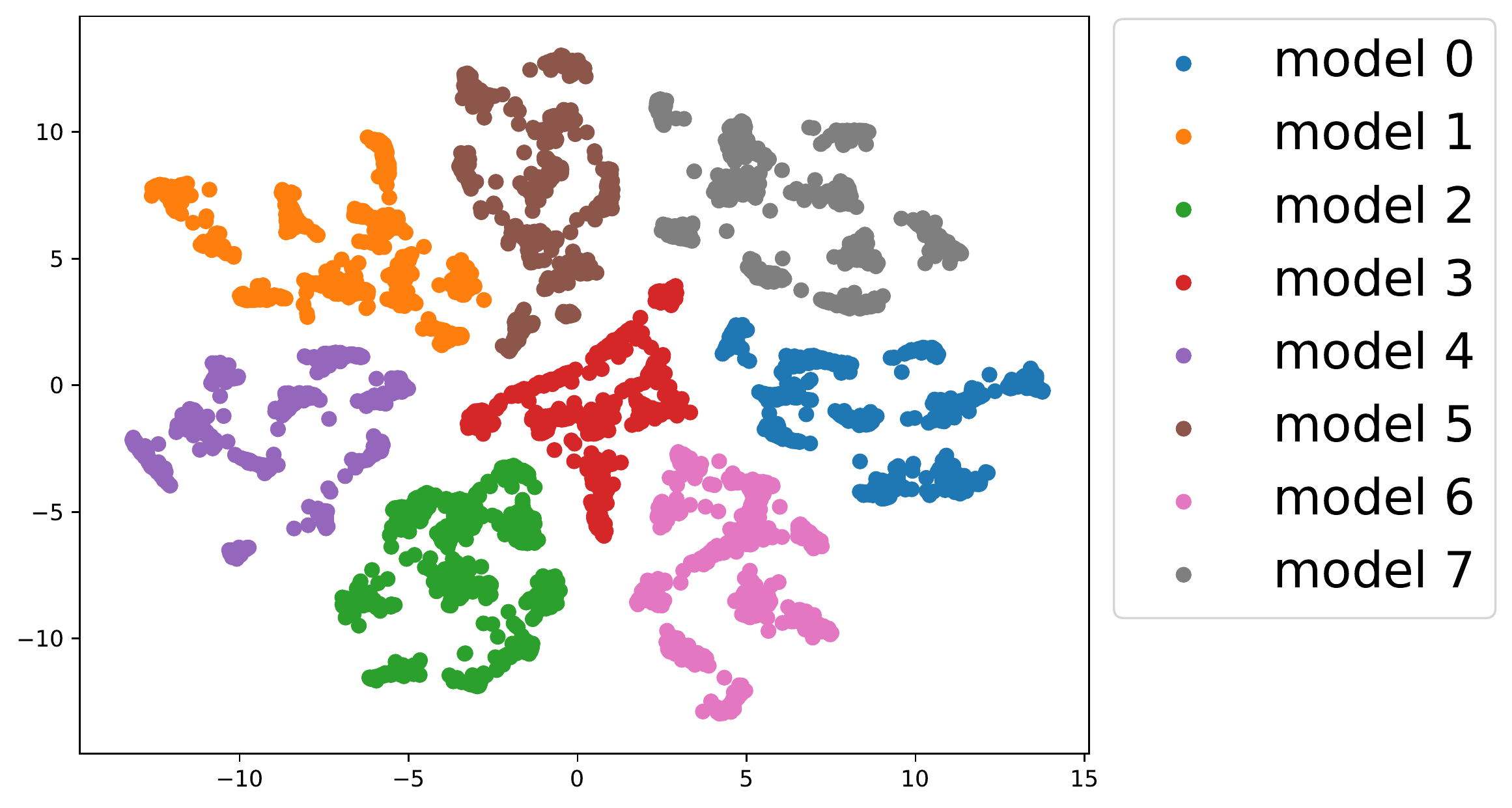}  
  \caption{Adversarially-disjoint (8 models).}
  \label{fig:ours_tsne4}
\end{subfigure}
\caption{A visualization using t-SNE embeddings of the adversarial gradients' signs for 1000 images for the adversarially-disjoint models, for a varying number of models in the set.}
  \label{fig:tsne}
\end{figure*}
\paragraph{Visualizing the gradients.}
Next, we visualize the separation of the adversarial gradients caused by our training scheme. Using the 3, 4, 7, and 8-models' set, we computed the signs of the adversarial gradients ($\nabla_xL_i$) for 1000 test images and we use the t-distributed stochastic neighbor embedding (t-SNE) method~\cite{van2008visualizing} to visualize them. We show this visualization in~\autoref{fig:tsne}. For comparison, we do the same process for 3 adversarially trained models in~\autoref{fig:tsne_at}. Adversarial training directly reduces the sensitivity of the model to the perturbations and it is not intended to separate the gradients of different models. However, we show it here to visualize the inherent or the baseline separation between different models' gradients (and therefore, to visualize the extent of separation done by our approach). As can be observed, in our approach, the models' gradients are well separated and distinctive from each other as intended by our transferability losses.

\begin{table*}[!t]
    \centering
    \setlength\tabcolsep{14pt}
    \resizebox{0.9\linewidth}{!}{
    \begin{tabular}{c|cc|ccccccc} \toprule
     \textbf{Attacks} & \multicolumn{2}{c|}{\textbf{Adversarially trained set}} & \multicolumn{6}{c}{\textbf{Adversarially-disjoint set (ours)}} \\ 
     & \textbf{3 models} & \textbf{4 models} & \textbf{3 models} & \textbf{4 models} & \textbf{5 models} & \textbf{6 models} & \textbf{7 models} & \textbf{8 models}\\ \midrule
     No attack & 85.2 & 85.3 & 94.1 & 94.0 & 93.7 & 93.5 & 93.8 & 93.5 \\ \midrule
     FGSM  & 60.5 & 61.1 & 66.5 & 73.5 & 78.6 & 81.6 & 83.9 & 85.4\\ 
     FGM & 45.6 & 46.2 & 66.2 & 72.1 & 78.0 & 81.2 & 83.5 & 85.1 \\ 
     R-FGSM & 73.3 & 73.7 & 65.5 & 72.5 & 76.5 & 79.3 & 81.7 & 82.8 \\ \midrule
     
     $\text{PGD}_1$ & 57.1 & 58.0 & 64.1 & 70.4 & 76.1 & 78.2 & 79.6 & 80.6 \\ 
     $\text{PGD}_2$ & 54.5 & 55.6 & 63.9 & 69.6 & 75.1 & 76.9 & 78.12 & 78.7 \\
     $\text{MI-FGSM}_1$  & 58.2 & 59.0 & 65.5 & 71.9 & 77.2 & 79.9 & 81.5 & 82.7 \\ 
     $\text{MI-FGSM}_2$ & 30.7 & 31.5 & 64.1 & 68.3 &73.5 & 73.8 &  73.2 & 73.2 \\ \midrule
    
     $\text{CW}_1$ & 60.1 & 63.6 & 62.7 & 69.9 & 74.8 & 77.9 & 80.3 & 81.6\\ 
     $\text{CW}_2$ & 7.0 & 7.2 & 59.7 & 63.2 & 63.7 & 65.3 & 71.1 & 69.2 \\
     $\text{EAD}_1$ & 39.5 & 44.3 & 59.2 & 66.1 & 71.0 & 75.1 & 77.8 & 79.4 \\ 
     $\text{EAD}_2$ & 12.5 & 12.6 & 53.8 & 58.8 & 66.6 & 67.9 & 77.4 & 72.5 \\ \bottomrule 
    \end{tabular}}
    \caption{Comparing our approach with adversarially trained sets in terms of test accuracy (\%). The first row shows the clean test accuracy (i.e. no attack). Other rows show the average accuracy of attacks over all combinations of source and target models in the set (i.e. including white-box and black-box cases). The attacks' parameters are in~\autoref{tab:attacks_params}.}
    \label{tab:set}
\end{table*}

\subsection{Robustness of the Set} \label{ref:whole_set}
In the last section, we investigated the transferability of attacks across the models and its relationship with increasing the number of models in the set. However, it is not very clear how much we gain when increasing the number of models, especially when the black-box accuracy declines for certain attacks. Additionally, it is not clear how our approach as a set compares to adversarial training, considering that adversarial training increases both white-box and black-box accuracy, while we only focus on black-box accuracy. Therefore, in this section, we evaluate the average accuracy of the whole set (i.e. the average of the full $n\times n$ attack matrix). In this evaluation, the baseline is also a set of adversarially trained models.  

We show our evaluation in~\autoref{tab:set} in which we report the whole-set average accuracy for the sets of 3 to 8 adversarially-disjoint models in comparison with the sets of 3 and 4 adversarially trained models. We also show the clean test accuracy (i.e. no attack) in the first row. 

We highlight the following observations from the table: 
\begin{enumerate}
    \item As reported in prior work~\cite{shafahi2019adversarial,madry2018towards,wong2019fast,tsipras2018robustness,xiao2020one}, adversarial training drops the accuracy of clean examples. On the contrary, our approach hardly drops the clean accuracy, even when training 8 models (the baseline clean accuracy is 94.13\%).
    \item With only 3 models in the set, our approach outperforms adversarial training in nearly all attacks (except R+FGSM). Adding only one more model (4 models) outperforms the adversarial training in all attacks. The other sets significantly outperform adversarial training. 
    \item Increasing the number of models in the adversarially trained set from 3 to 4 has very little performance gain in most of the cases. On the other hand, increasing the number of models in our approach can increase the performance by a large extent (e.g. when comparing the 4-models' set to the 3-models' set).
    \item In some cases, there is no significant performance gain when adding a new model (e.g. from 7 to 8 models), since the black-box accuracy can drop in some cases (see~\autoref{tab:transerability}). However, adding a new model (e.g. an 8th model) had a comparable performance and in most cases did not harm the average accuracy. Additionally, it has other advantages in the advanced attacks when the attacker has access to more than one model (discussed in the next section). 
\end{enumerate}

\paragraph{Expanding to higher perturbations budgets.}
In order to test the extent of our defense, we examine strong PGD attacks with significantly high perturbations' budgets and a large number of steps. We vary $\epsilon$ from 0.0196 to 0.156 and for each value, we run the PGD attack with 100 steps and $\alpha=2.5\cdot\epsilon/100$ (following~\cite{madry2018towards} for comparison). We run this experiment on the 5 and 7-models' set as a representative since the black-box accuracy tends to decline with increasing the number of models. For comparison, we also perform the attack on a 5-models' adversarially trained set. 

\begin{figure} [!t]
    \centering
    \includegraphics[width=\linewidth]{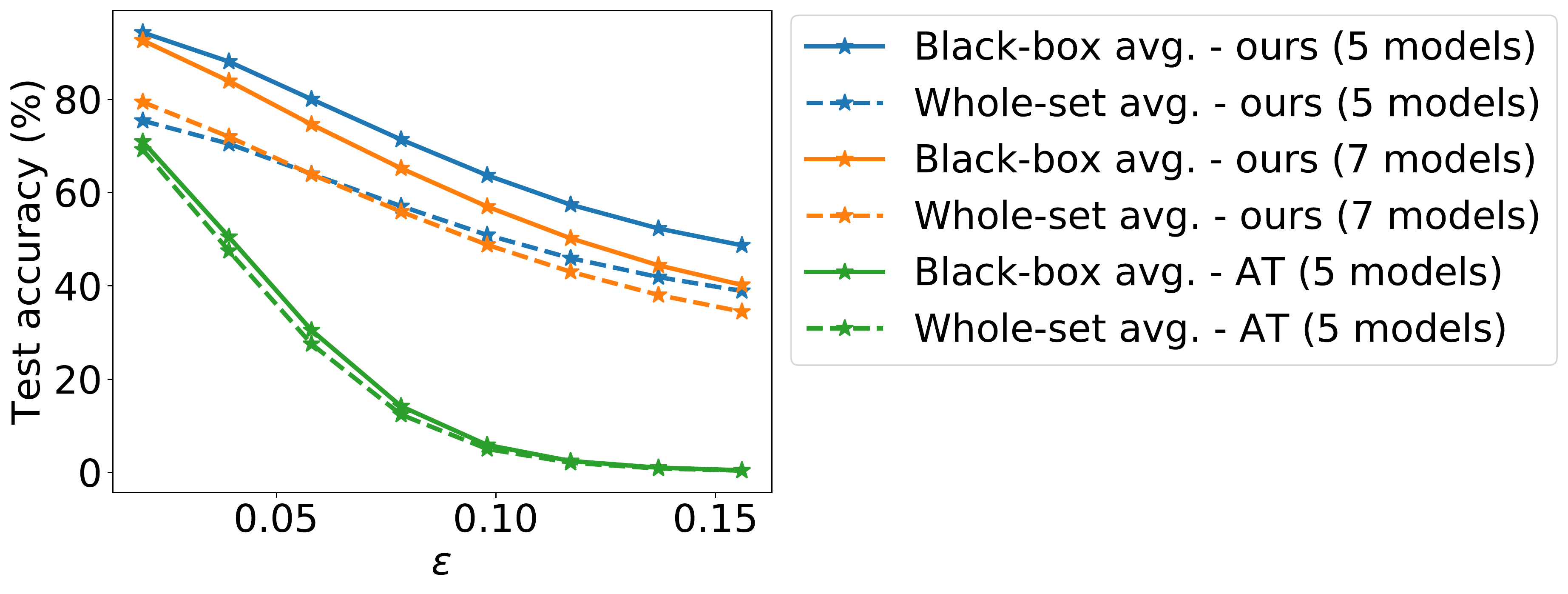}
    \caption{The average accuracy of black-box attacks across the models and the accuracy of the whole set with increasing values of the perturbation budget ($\epsilon$). We run the attack on the 5 and 7 adversarially-disjoint models in comparison with 5 adversarially trained models. The attack implemented is PGD with 100 steps and $\alpha = 2.5\cdot\epsilon/100$~\cite{madry2018towards}.}
    \label{fig:pgd_range}
\end{figure}

We show the average black-box accuracy across the sets in~\autoref{fig:pgd_range} in addition to the whole-set average. From the figure, we can observe that the black-box accuracy decreases, especially for larger perturbations, in the case of the 7 models compared to the 5 models. However, the whole-set average is roughly comparable, with the 7-models' average being slightly higher for smaller perturbations and the 5 models' average being slightly higher for larger ones. 

However, for both sets, the black-box and whole-set average numbers are significantly higher than the adversarially trained set. Even with such strong perturbations that significantly reduce the effectiveness of PGD adversarial training~\cite{madry2018towards}, and at $\epsilon=0.117$, our approach still has an average accuracy of 45.94\% and 43.02\% for the 5 and 7 sets respectively. Interestingly, this is roughly on par with the white-box robustness of one adversarially trained model using a much weaker PGD attack with $\epsilon=0.031$ and 20 steps~\cite{madry2018towards}. 

\subsection{Attacks using Multiple Models}
Our standard assumption is that the adversary can access only one model from the set. In this section, we extend this assumption and evaluate against an `oracle adversary' having white-box access to more models (e.g. by purchasing many versions that could randomly have different models).  

\paragraph{Attacks using an ensemble.} If the adversary can access $m$ models (where $m \leq n$), they can fuse their logits to form a new model: $$f_\text{ens} = \frac{1}{m}\sum_{i=0}^{m}f_i$$ 
which is used to compute the loss: $L_\text{ens}$. One can now form the adversarial image using the adversarial gradient of $L_\text{ens}$ (i.e. $\nabla_xL_\text{ens}$). We use this setup in our following evaluations. 

We also experimented with another approach that averages the gradients of the models ($\frac{1}{m}\sum_{i=0}^{m}\nabla_xL_m$) and use the average in the attacks, however, it was not very effective compared to using $\nabla_xL_\text{ens}$. 

\begin{table}[!t]
    \centering
    \resizebox{0.65\linewidth}{!}{
    \begin{tabular}{c|cccc} \toprule
     \textbf{Source model(s)} & \multicolumn{4}{c}{\textbf{Target model}} \\
     & \textbf{0} & \textbf{1} & \textbf{2} & \textbf{3} \\  \midrule 
      0 &  0 & 95.1 & 95.8 & 89.9 \\
      1 & 90.9 & 0.0 & 92.2 & 88.8 \\
      2 & 88.8 & 94.7 & 0.0 & 92.5 \\ 
      3 & 95.1 & 95.4 & 94.1  & 0.0 \\ \midrule
      0,1 & 0.0 &  0.0 & 95.9 & 90.8 \\
      0,2 & 0.0 & 96.2 & 0.0 & 94.3 \\
      0,3 & 0.0 & 96.5 & 97.2 & 0.0 \\ \midrule
      0,1,2 & 0.0 & 0.0 & 0.0 & 96.9 \\
      0,2,3 & 0.0 & 98.4 & 0.0 & 0.1 \\ \midrule 
      0,1,2,3 & 0.0 & 0.1 & 0.1 & 0.1 \\ \bottomrule
    \end{tabular}}
    \caption{Test accuracy (\%) of ensemble-based attacks against the 4-models set. Attacks are computed using an ensemble of an increasing number of models (1, ..., n) and evaluated on all models in the set. The attack implemented is PGD with 20 steps ($\epsilon=0.031$, $\alpha=0.0078$).}
    \label{tab:ens_4models}
\end{table}

\paragraph{Ensemble attacks against the 4-models set.} We used the previously mentioned approach to attack the 4-models' set. We show this experiment in~\autoref{tab:ens_4models}. We vary the number of models on which we compute the attacks: from 1 model (which is similar to all previous experiments) to 4 (using all models in the set). We then use the crafted images to attack each model individually (since these models are individually deployed as per our assumption). The attack used in this experiment is PGD with $\epsilon=0.031$, $\alpha=0.0078$, and 20 steps ($\text{PGD}_2$ in the previous tables). We give examples of different combinations of source models in~\autoref{tab:ens_4models}. We can observe that when a model $f_i$ is included in the ensemble, the accuracy of $f_i$ on the crafted image will drop to nearly 0 (similar to the white-box case for each model individually). Therefore, accessing all models drops the accuracy to nearly 0 on all models. However, if a model is not included in the ensemble, its accuracy on the attack images is still significantly high. 

\paragraph{Attacks versus the number of models in the set.} The previous experiment shows that the success of the attack depends on how many models the adversary can access out of the whole set. So, a natural next step is to evaluate the attacks w.r.t the number of models in the set.

\begin{figure} [!t]
    \centering
    \includegraphics[width=0.8\linewidth]{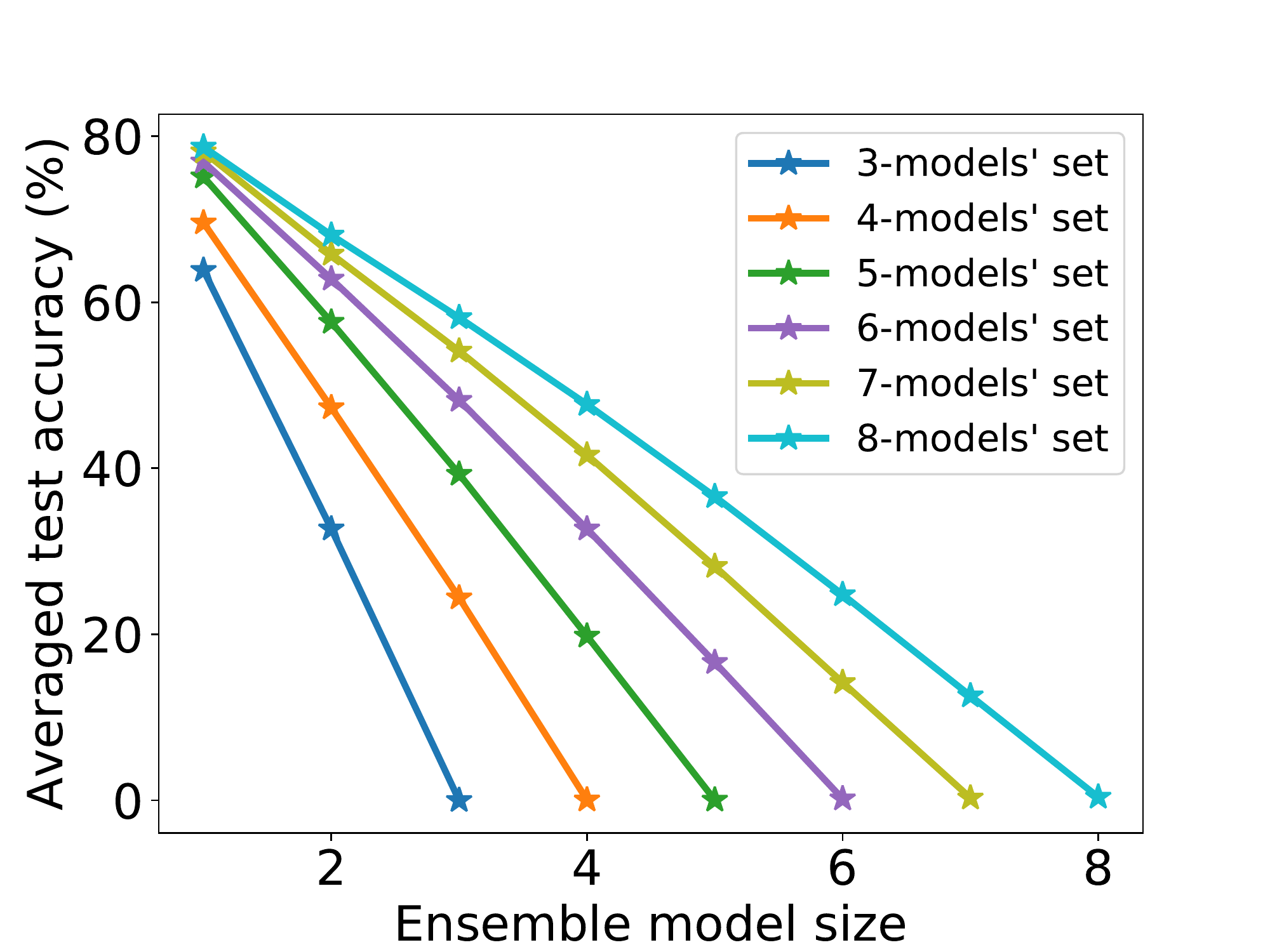}
    \caption{Test accuracy (averaged over all models in the set) of attacks crafted using an ensemble with a varying number of models (x-axis). The analysis is repeated for all sets. The attack is PGD with 20 steps ($\epsilon=0.031$, $\alpha=0.0078$).}
    \label{fig:ens_attacks}
\end{figure}

Thus, we extend the previous analysis done in~\autoref{tab:ens_4models} to all sets, using the same setup (i.e. the attacks are crafted on the ensemble and tested on all models). We show in~\autoref{fig:ens_attacks} the relationship between the number of models in the attacker's ensemble and the average accuracy on all models in the set. Each point represents an average over all combinations of $\Comb{n}{m}$, where $n$ is the total number of models in the set, and $m$ is the number of models in the ensemble. From this figure, we observe that having a larger number of models in the set helps to alleviate this attack since we make it harder for the adversary to acquire all the models. Therefore, even when there is no significant gain in the whole-set average (e.g. such as when increasing from 7 to 8 models in~\autoref{tab:set}), it is still beneficial to increase the number of models (to potentially beyond 8 models) to mitigate advanced attacks.   

\subsection{Ablation Study}
In this section, we present an analysis of some design choices and components in our approach. 

\paragraph{Angular deviation and transferability losses.} In our approach, we used both angular deviation loss and transferability losses. We here investigate each component individually. 
\begin{table}[!b]
    \centering
    \resizebox{0.65\linewidth}{!}{
    \begin{tabular}{c|c} \toprule 
     \textbf{Training variant} & \textbf{Black-box accuracy} \\ \midrule
     No transferability & 21\% \\
     No angular deviation & 85.9\%\\
     Both losses & 96.2\% \\ \bottomrule
    \end{tabular}}
    \caption{A comparison between three variants in terms of black-box accuracy: training with angular deviation loss only, training with transferability loss only, and training with both losses. The attack is $\text{PGD}_1$ in~\autoref{tab:attacks_params}. The number of models in the set is 3.}
    \label{tab:ablation1}
\end{table}
Intuitively, increasing the angular deviation between two gradients should help to make their losses uncorrelated. To test this, we trained 3 models using the classification loss and the angular deviation loss only. We evaluated the performance of PGD attack on a validation set during training. We found that increasing the angular deviation does help to reduce the transferability in the first few epochs (average black-box attack is 30-40\% after few epochs). However, the black-box accuracy decreases again during training even when the angular deviation loss is also decreasing. 

As shown in~\autoref{tab:ablation1}, at the end of training, the black-box accuracy decreased to 21\%. This suggests that angular deviation (on its own) is not enough to achieve low transferability. This is also supported by previous work~\cite{liu2017delving} that showed that two models with orthogonal gradients can still have high transferability of attacks. Our interpretation is that the hypothesis that orthogonal gradients lead to low transferability is driven by the linear approximation of the loss function. Since, in fact, the loss function is not linear, there might be two orthogonal directions that both increase the loss of one model. 

On the other hand, we trained another variant with transferability losses only without angular deviation, which we show in the second row of~\autoref{tab:ablation1}. This variant is much more successful in decreasing the transferability than the first one. However, what worked best which we eventually used, is to train with both losses at first and then after few epochs ($\sim$8 epochs), stop the angular deviation and train with transferability losses only. Our interpretation is that diversifying the models' gradients with angular deviation helps to find a good starting point to search for the adversarially disjoint gradients by the transferability losses.

\begin{table}[!b]
    \centering
    \resizebox{0.85\linewidth}{!}{
    \vspace{-3mm}
    \begin{tabular}{cc|cc} \toprule
     \textbf{Training variant} & \textbf{models} & \textbf{Black-box (\%)} & \textbf{Clean (\%)} \\ \midrule 
     Joint & 3 & 95.7 & 94.1 \\
     Joint & 4 & 93.8 & 94.0 \\ \midrule
     Joint & 5 & 87.9 & 91.3 \\
     add & 5 & 88.7 & 90.5 \\ 
     Sampling & 5 & 95.12 & 93.7 \\  \bottomrule
    \end{tabular}}
    \caption{A comparison in terms of accuracy between expanding the number of models using joint training, adding one model while fixing the others, and joint training with sampling. The attack is $\text{PGD}_1$ in~\autoref{tab:attacks_params}.}
    \label{tab:ablation2}
\end{table}

\paragraph{Increasing the number of models.}
Jointly training more than 5 models with all combinations of transferability losses at each mini-batch was not very successful; the losses were fluctuating and the training converged to both lower classification accuracy and lower black-box accuracy than the 3 and 4 models cases. In addition, it required additional training time and memory resources. This could possibly be due to the difficulty of jointly optimizing all the losses simultaneously.  

To circumvent this, we experimented with adding one new fifth model while fixing the weights of the trained four models. However, although faster, this was also not very effective. This is probably because our approach is more effective when each model changes its own adversarial gradients. Also, the new model's clean accuracy was slightly lower than the first few four models. 

Finally, what worked best in our setup, is to jointly train models but use only a few randomly sampled models each iteration for the transferability loss. This stabilized the training without any significant fluctuation and significantly reduced the time each epoch takes because even when we have a larger number of models in the set, we compute the expensive backward pass for only 3 models. Also, the clean accuracy was close to the 3 and 4 models case. 
We compare these three approaches in~\autoref{tab:ablation2} in terms of clean accuracy and black-box accuracy (using $\text{PGD}_1$ attack in~\autoref{tab:attacks_params}). 

\section{Discussion}
\new{In this section, we discuss the limitations, implications, and possible extensions for our work.}
\new{\paragraph{Number of models.} In this work, we take the first step towards randomized deployment defenses for adversarial robustness. Using our approach, we managed to expand the number of models to 8 models without affecting the clean accuracy, but with a slight drop in the black-box accuracy. However, since increasing the number of models is important for mitigating advanced attacks and for improving the whole-set average, it still remains an open issue of how to, efficiently and with maintaining the performance, train and generate a larger population of models at scale and also how to increment the number of models without re-training.}

\new{\paragraph{White-box robustness.} The `adversarially-disjoint' models have significantly lower attack transferability, however, they have no white-box robustness. Even though we manage to (often significantly) outperform adversarial training in many settings (see~\autoref{fig:pgd_range} and~\autoref{tab:set}), it is a natural extension to investigate whether it is possible to combine the adversarial robustness with disjointing, or whether they are at odds as a fundamental trade-off (e.g. similar to the trade-off between robustness and accuracy~\cite{tsipras2018robustness}).}

\section{Conclusion}
In this work, we propose a new paradigm for adversarial defenses. Instead of attempting to solve the intrinsic vulnerabilities of DNNs, we exploit the unexplored opportunities of models' deployment. We propose to change the traditional white-box threat model by deploying same-functionality, yet, adversarially-disjoint models with minimal transferability of attacks. By doing so, even if the adversary can craft successful adversarial attacks on one model, they are significantly less successful on other deployed models. Our approach is fundamentally different from other randomness defenses as it does not involve any gradient masking or obfuscation that could be easily circumvented.

To obtain the adversarially-disjoint models, we propose a novel gradient penalty that strongly reduces the transferability of attacks across the models. Our approach is significantly more resilient to transferability in comparison to a baseline of ensemble diversity. Additionally, we outperform a baseline of an adversarially trained set over a wide range of attacks, while hardly having any negative effect on the clean accuracy. 

\section{Acknowledgement}
We thank Yang Zhang and Hossein Hajipour for constructive advice and valuable discussions.

\printbibliography
\end{document}